%% file: main.tex
\documentclass{article}

\usepackage[dblblindworkshop, final]{neurips_2026}
\setcitestyle{numbers}
\workshoptitle{Representations for the Physical Sciences}

\usepackage[utf8]{inputenc}
\usepackage[T1]{fontenc}
\usepackage{hyperref}
\hypersetup{hidelinks}
\usepackage{url}
\usepackage{booktabs}
\usepackage{amsmath}
\usepackage{amsfonts}
\usepackage{amssymb}
\usepackage{nicefrac}
\usepackage{microtype}
\usepackage{xcolor}
\usepackage{graphicx}
\usepackage{enumitem}
\usepackage{tikz, pgfplots}
\usepackage{wrapfig}
\usetikzlibrary{arrows.meta,positioning,fit,calc,shapes.geometric}

\title{Does Text Steer Neural PDE Surrogates?\\A Controlled Diagnostic with OperatorCLIP}

\author{
  Aadi Dash$^{1}$, Lennon J. Shikhman$^{1,2,*}$, Michael Galarnyk$^{3}$ \\
  $^{1}$College of Computing, Georgia Institute of Technology\\
  $^{2}$Department of Mathematics and Systems Engineering, Florida Institute of Technology\\
  $^{3}$Wallace H. Coulter Department of Biomedical Engineering, Georgia Institute of Technology\\$^{*}$Corresponding author\\
  \texttt{\{aadidash, lshikhman3, mgalarnyk3\}@gatech.edu}
}

\begin{document}
\maketitle
\begingroup
\renewcommand{\thefootnote}{}
\endgroup

\begin{abstract}
Lower error from a text-conditioned neural surrogate does not, by itself, show that the model uses the meaning of the text. We examine this attribution problem with \emph{OperatorCLIP}, comparing an unconditioned FNO, a constant-sentence FiLM control, and a fixed task description trained with contrastive alignment. Three-seed experiments cover Darcy2D, ShallowWater2D, and three-dimensional compressible Navier--Stokes (CNS3D). Constant conditioning has lower mean test error on both 2D tasks. Relative to this control, task text plus alignment has a similar mean on ShallowWater2D and CNS3D and a higher mean on Darcy2D; these descriptive comparisons have substantial seed uncertainty. The latter comparison changes both prompt content and loss, so it isolates neither effect. The text encoder is trained from scratch, and each conditioned model sees only one description during training. In this regime, pairwise InfoNCE cannot identify matched pairs and has minimum $\log B$. Prompt interventions show no reliable semantic ordering. This methodological caution demonstrates why pathway controls are needed; it neither establishes semantic competence of the encoder nor tests the effectiveness of text under varying physical context.
\end{abstract}

\section{Introduction}
\label{sec:introduction}

Neural PDE surrogates approximate solution operators and time-advance maps \cite{kovachki2023neuraloperator}. Observed fields may underdetermine behavior when coefficients \cite{li2021fourier}, boundary conditions \cite{shikhman2026one}, forcing or operator examples \cite{yang2023incontext}, geometry \cite{li2023geofno}, or discretization \cite{gao2025discretization} vary.

PDEBench exposes such context through varying physical and simulation settings \cite{takamoto2022pdebench, shikhman2026diagnosing}. Context may also appear in metadata, solver configurations, or scientific language, motivating models conditioned on text, equations, and operator examples \cite{lorsung2024explain,liu2024prosefd,yang2023incontext}.

Adapters, affine modulation, extra parameters, and auxiliary losses can alter optimization even when a prompt is constant. A task-specific model can therefore improve without using linguistic content. Establishing semantic steering requires matched pathway controls and prompt interventions, beyond comparison with an unconditioned model.

OperatorCLIP combines CLIP-style alignment \cite{radford2021clip} with FiLM under the same FNO backbone specification. Constant-prompt controls assess pathway and capacity changes together. Correct, missing, shuffled, corrupted, paraphrased, and numeric-metadata prompts probe sensitivity after training. Since each model trains on one description, these interventions cannot establish that it learned how changing physical context should change predictions.

\paragraph{Contributions.}
We document attribution limits of fixed-prompt comparisons, prove that identical deterministic text embeddings make pairwise alignment unidentifiable, and report all 27 runs across two 2D tasks and one 3D task. The controls expose confounds rather than fully separating their causal effects.

\section{Related Work}
\label{sec:related_work}

\paragraph{Operators and context.}
Operator architectures include DeepONet \cite{lu2021deeponet}, spectral FNO \cite{li2021fourier}, geometry-aware models \cite{li2023geofno,chen2026argent}, CNO and U-Net \cite{raonic2023convolutional,ronneberger2015unet}, and attention-based OFormer and GNOT \cite{li2023oformer,hao2023gnot}. Physics-informed networks impose equations through the loss \cite{raissi2019pinn}; meta-learning shares information across parametric families \cite{cho2023hyperlrpinn,chen2024gptpinn}. All our comparisons use FNO.

\paragraph{Benchmarks and failure modes.}
PDEBench spans eight PDE families in one to three dimensions \cite{takamoto2022pdebench}; PDEArena varies parameters and scales \cite{gupta2023pdearena}; and CFDBench varies boundaries and geometry \cite{luo2023cfdbench}. The Well, BubbleML, and APEBench add further systems and rollout settings \cite{ohana2024well,hassan2023bubbleml,koehler2024apebench}. Audits show dependence on boundary distributions \cite{shikhman2026one}, failure under parameter or boundary shifts \cite{shikhman2026diagnosing}, discretization mismatch error \cite{gao2025discretization}, and inconsistency under temporal composition \cite{shikhman2026semigroup}. These settings motivate testing whether text supplies information beyond the numerical input.

\paragraph{Multimodal scientific learning.}
PROSE and PROSE-FD jointly process equations and fields
\cite{liu2024prose,liu2024prosefd}; language-conditioned surrogates use cross-attention \cite{lorsung2024explain}; and in-context operators accept data-pair prompts \cite{yang2023incontext}, with multimodal
extensions incorporating equations and natural-language descriptions \cite{yang2023multimodal}. Unisolver and PDEformer encode PDE components or graphs \cite{zhou2025unisolver,ye2024pdeformer}, while DPOT, Poseidon, and MPP pretrain across systems without text \cite{hao2024dpot,herde2024poseidon,mccabe2024mpp}. OperatorCLIP combines contrastive alignment \cite{radford2021clip} with FiLM \cite{perez2018film}; PICL derives similarity from equation coefficients \cite{lorsung2024picl}; and OmniArch aligns captions with spectral changes \cite{chen2025omniarch}. Our diagnostic examines attribution limits of fixed-text conditioning.

\section{Controlled Diagnostic Framework}
\label{sec:operatorclip}

\begin{wrapfigure}{r}{0.51\textwidth}
\vspace{-1.2em}
\centering
\resizebox{\linewidth}{!}{%
\begin{tikzpicture}[
  font=\scriptsize,
  >={Latex[length=1.5mm]},
  block/.style={draw=black!55, rounded corners=1pt, fill=white,
                minimum height=4.2mm, inner xsep=3pt, inner ysep=1.5pt},
  prompt/.style={block, fill=blue!7},
  model/.style={block, fill=black!5, minimum width=16mm},
  flow/.style={->, line width=0.45pt, draw=black!75}
]
  \node[font=\scriptsize\bfseries, anchor=east] at (1.55,2.55) {Unconditioned};
  \node[block] (x0) at (1.95,2.55) {$x$};
  \node[model] (m0) at (3.95,2.55) {FNO};
  \node[block] (y0) at (5.05,2.55) {$\widehat y$};
  \draw[flow] (x0) -- (m0);
  \draw[flow] (m0) -- (y0);

  \node[font=\scriptsize\bfseries, anchor=east] at (1.55,1.45) {Constant};
  \node[block] (x1) at (1.95,1.45) {$x$};
  \node[prompt] (p1) at (2.95,1.88) {generic sentence};
  \node[model] (m1) at (3.95,1.45) {FiLM--FNO};
  \node[block] (y1) at (5.05,1.45) {$\widehat y$};
  \draw[flow] (x1) -- (m1);
  \draw[flow] (p1) -- (m1);
  \draw[flow] (m1) -- (y1);

  \node[font=\scriptsize\bfseries, anchor=east] at (1.55,0.35) {Aligned};
  \node[block] (x2) at (1.95,0.35) {$x$};
  \node[prompt] (p2) at (2.95,0.78) {fixed task text};
  \node[model] (m2) at (3.95,0.35) {FiLM--FNO};
  \node[block] (y2) at (5.05,0.35) {$\widehat y$};
  \draw[flow] (x2) -- (m2);
  \draw[flow] (p2) -- (m2);
  \draw[flow] (m2) -- (y2);
  \node[block, fill=orange!9, font=\tiny] (nce) at (2.95,-0.08) {InfoNCE};
  \draw[flow] (nce) -- (p2);

  \node[block, fill=black!2, font=\tiny, minimum width=50mm] (tests) at (3.25,-0.72)
    {Test-time: correct $\mid$ corrupted $\mid$ paraphrased $\mid$ missing};
\end{tikzpicture}%
}
\caption{Diagnostic comparisons. Unconditioned-to-constant changes pathway and capacity together; constant-to-aligned changes both text and loss. Interventions probe prompt sensitivity.}
\label{fig:operatorclip_overview}
\vspace{0.5em}
\end{wrapfigure}
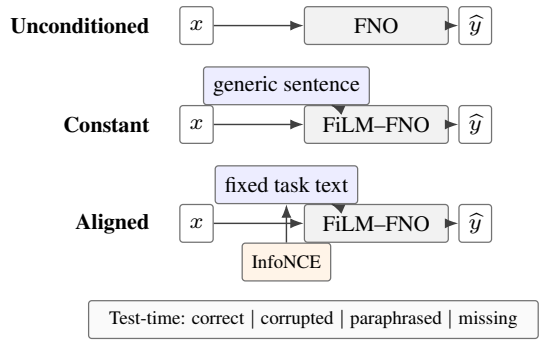

For input field or history \(x_i\) and target \(y_i\), an unconditioned surrogate predicts \(\mathcal{G}_\theta(x_i)\). OperatorCLIP encodes operator context \(m_i\), such as coefficients, boundaries, or geometry, in a structured prompt \(p_i\) and predicts
\begin{equation}
    \widehat{y}_i = \mathcal{G}_\theta(x_i; h_i^{\mathrm{text}}),
\end{equation}
where \(h_i^{\mathrm{text}}\) is the text encoder's conditioning vector. The projection heads below are used only for the auxiliary alignment loss.

\paragraph{Shared text-data embedding.}
A text encoder maps \(p_i\) to \(h_i^{\mathrm{text}}=E_{\mathrm{text}}(p_i)\), while a data encoder maps numerical evidence to \(h_i^{\mathrm{data}}=E_{\mathrm{data}}(x_i,y_i)\). For \(q\in\{\mathrm{text},\mathrm{data}\}\), projection heads produce normalized embeddings
\begin{equation}
    z_i^q = \frac{P_q h_i^q}{\|P_q h_i^q\|_2}.
\end{equation}
For a minibatch of \(B\) matched text-data pairs, with \(S_{ij}=\langle z_i^{\mathrm{text}},z_j^{\mathrm{data}}\rangle/\tau\), we use the symmetric contrastive loss
\begin{equation}
\mathcal{L}_{\mathrm{align}}
=
\frac{1}{2B}
\sum_{i=1}^{B}
\left[
-\log\frac{\exp(S_{ii})}{\sum_j\exp(S_{ij})}
-\log\frac{\exp(S_{ii})}{\sum_j\exp(S_{ji})}
\right].
\end{equation}
The training objective is
\begin{equation}
    \mathcal{L}
    =
    \mathcal{L}_{\mathrm{pred}}
    +
    \lambda_{\mathrm{align}}\mathcal{L}_{\mathrm{align}},
\end{equation}
where \(\mathcal{L}_{\mathrm{pred}}\) is the supervised field or trajectory loss.

\paragraph{Proposition 1 (fixed-prompt non-identifiability).}
\emph{If every example produces the same deterministic text embedding, then $\mathcal{L}_{\mathrm{align}}\geq\log B$ and the loss contains no information identifying matched text--data pairs.}
In the data-to-text direction, all $B$ candidates are identical and the loss is exactly $\log B$. In the text-to-data direction, every row shares one score distribution while each data item serves once as the positive; the batch average is minimized by a uniform distribution, again at $\log B$. Away from equal scores, alignment can still produce gradients; the result does not imply a constant loss throughout training or fix the learned embedding. Shuffling identical prompts cannot change the model inputs.

\paragraph{Text encoder and conditioning.}
Both conditioned configurations use a text encoder trained jointly with the FNO from random initialization, without language pretraining. Its output supplies the FiLM conditioning vector. Appendix~\ref{sec:run_details} specifies the architecture, tokenizer, and initialization.

\section{Results}
\label{sec:results}

\paragraph{Controlled setup.}
Each FNO is trained separately on one task. Darcy2D maps a $128\!\times\!128$ coefficient field to its solution using $8{,}000/1{,}000/1{,}000$ train/validation/test examples; ShallowWater2D maps the first $128\!\times\!128$ state to the next using an $800/100/100$ split. CNS3D maps five-channel $128^3$ states one step forward using $80/10/10$ examples. The 2D FNO has width 64 and 12 modes per axis and trains for 50 epochs with batch size four; the 3D FNO has width 48 and eight modes per axis and trains for 30 epochs with batch size two. All use four layers, channel normalization, AdamW ($5\!\times\!10^{-4}$), optimization seeds 7, 17, and 27, and the same fixed data split. We report final-epoch test relative $L^2$ as mean $\pm$ sample standard deviation, with every seed in Appendix~\ref{sec:run_details}.

The \emph{unconditioned} FNO removes the text encoder and FiLM modules. The \emph{constant} control retains them but always receives ``Operator context unavailable. Predict the output from the numerical input,'' with no alignment loss. The \emph{aligned} condition uses a fixed task description listing dataset, dimension, channels, task type, input/target steps, filename, and data key, with $\lambda_{\rm align}=0.1$ and temperature $0.07$. Its numerical-evidence encoder uses per-channel input/target statistics. Targets enter only the auxiliary training loss, never inference. Conditioning adds approximately $2.8$--$2.9\%$ surrogate parameters in 2D and $0.6\%$ in 3D. Constant-to-aligned changes the prompt, vocabulary size, and objective together; it is not an isolated alignment ablation.

For semantic steering, prompt dependence should be accompanied by an ordering consistent with meaning: correct and meaning-preserving descriptions should outperform missing or factually corrupted ones. Here that criterion is a diagnostic, not a capability learned under varying training prompts.

\begin{table}[h]
\centering
\small
\setlength{\tabcolsep}{3.5pt}
\caption{Test relative $L^2$ (mean $\pm$ sample s.d., three seeds). Positive $\Delta_{\rm U\to C}$ and $\Delta_{\rm C\to A}$ denote lower mean error in constant versus unconditioned and aligned versus constant. Percentages are descriptive ratios of means, not isolated causal effects or significance claims.}
\label{tab:controlled_results}
\input{generated/controlled_results}
\end{table}

The constant mean is lower than the unconditioned mean on both 2D tasks (Table~\ref{tab:controlled_results}). On ShallowWater2D, the aligned mean is only $1.5\%$ lower than the constant mean; the $18.9\%$ difference from unconditioned would conflate this small difference with the pathway/capacity change. Darcy2D favors constant in the observed means, while CNS3D means differ by less than $0.6\%$. Three seeds on one split do not establish stable rankings or variance effects. Alignment losses approach $\log B$ ($1.386$ for $B=4$ in 2D and $0.693$ for $B=2$ in 3D), as permitted by Proposition~1.

\begin{figure}[t]
\centering
\includegraphics[width=\linewidth]{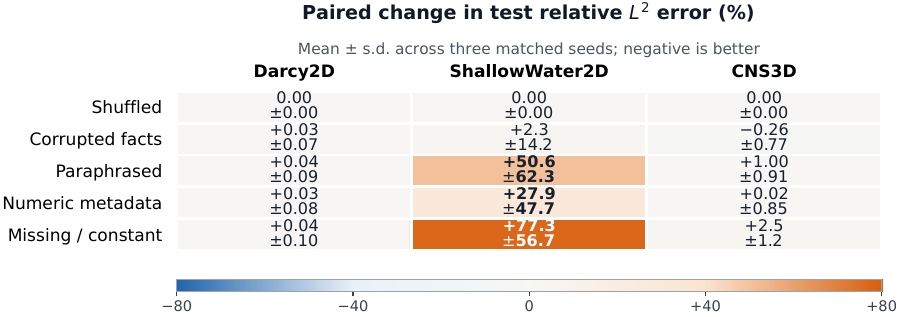}
\caption{Prompt sensitivity of the aligned models. Cells show the paired change in test relative $L^2$ error from the correct description (mean $\pm$ sample standard deviation over the same three seeds); positive values indicate higher error. Missing and constant modes use the same sentence.}
\label{fig:prompt_controls}
\end{figure}

The interventions do not establish semantic ordering (Figure~\ref{fig:prompt_controls}). Shuffling is exactly neutral because prompts are identical within each task. Darcy2D is nearly invariant; CNS3D changes are smaller in magnitude than ShallowWater2D's. On ShallowWater2D, corrupted facts yield a paired error change of $+2.3\%\pm14.2\%$, compared with $+50.6\%\pm62.3\%$ for paraphrases. These variable responses show embedding sensitivity without establishing a semantic explanation. Because these alternative descriptions are unseen in training, their effects also depend on tokenization and representation quality.

\section{Discussion}
\label{sec:discussion}

The constant control shows that lower observed mean error can accompany a fixed generic prompt. Its embedding drives a learned, dataset-wide affine modulation, but the experiment does not separate FiLM, added capacity, and optimization effects. Capacity-matched FNOs and learned task identifiers remain necessary controls. Differences in dataset size, dimension, and training budget also prevent attributing cross-task patterns to PDE family alone.

The encoder has no pretrained semantic competence, and one-description training supplies no supervision for how changes in text should affect predictions. We have not assessed representation quality independently, including distances between intervention embeddings. These experiments cannot distinguish inadequate representations, representations the surrogate has not learned to use, and context already implicit in a task-specific model. The expected non-identifiability under this design is a methodological caution, not evidence against text conditioning generally.

The aligned runs use a fixed $\lambda_{\rm align}=0.1$; no selection or sensitivity study is reported. A same-task-prompt control with $\lambda_{\rm align}=0$, followed by a weight sweep, is needed to isolate alignment and assess robustness. Proposition~1 holds independently of that weight, but empirical rankings need not. A natural next experiment jointly trains on varying coefficients, boundaries, forcing, geometry, or horizons, with held-out contexts, exact numerical metadata, and representation checks. More seeds and data splits would improve uncertainty assessment.

\section{Conclusion}
\label{sec:conclusion}

The constant-prompt results and fixed-prompt non-identifiability result show why improvement over an unconditioned baseline is insufficient evidence that text meaning helped. The observed means and prompt interventions do not isolate the causes of performance differences or establish semantic steering. Testing that capability requires varying physical context during training, matched capacity and loss controls, and evidence that the text representation preserves the relevant meaning.

\newpage
\section*{Acknowledgements} \paragraph{Funding.} This work was supported in part by Dell Technologies. Lennon J. Shikhman and Michael Galarnyk acknowledge financial support from Dell Technologies. Computational resources provided by Dell Technologies and used in this work include a Dell Pro Max T2 workstation equipped with an Intel Core Ultra 9 285K processor, 128 GB of DDR5 ECC memory, and an NVIDIA RTX PRO 6000 Blackwell GPU, and a Dell Pro Max 16 Plus laptop equipped with an Intel Core Ultra 9 285HX processor, 128 GB of DDR5-6400 memory, and an NVIDIA RTX PRO 5000 Blackwell GPU.

\bibliographystyle{plainnat}
\begingroup
\small
\raggedright
\bibliography{references}
\endgroup
\clearpage
\appendix
\input{run_details}
\end{document}

%% file: generated/controlled_results.tex
\begin{tabular}{lccccc}
\toprule
Task & Unconditioned & Constant & Fixed prompt + align. & $\Delta_{\rm U\to C}$ & $\Delta_{\rm C\to A}$ \\
\midrule
Darcy2D & $.03714\!\pm\!.00744$ & $.03009\!\pm\!.00064$ & $.03903\!\pm\!.00941$ & $+19.0\%$ & $-29.7\%$ \\
ShallowWater2D & $.001694\!\pm\!.000242$ & $.001394\!\pm\!.000272$ & $.001373\!\pm\!.000099$ & $+17.7\%$ & $+1.5\%$ \\
CNS3D & $.40446\!\pm\!.00485$ & $.40675\!\pm\!.00789$ & $.40458\!\pm\!.01458$ & $-0.6\%$ & $+0.5\%$ \\
\bottomrule
\end{tabular}

%% file: run_details.tex
\section{Run-level results and implementation details}
\label{sec:run_details}

Table~\ref{tab:seed_results} gives every final-epoch test error on the fixed split. Matching seed labels do not guarantee identical backbone initializations or minibatch orders because model construction changes random-number consumption. The runs measure optimization variation, not uncertainty over splits or PDE regimes. Table~\ref{tab:controlled_results} reports sample standard deviations, not confidence intervals. Its percentages are $100(1-\overline e_{\rm C}/\overline e_{\rm U})$ and $100(1-\overline e_{\rm A}/\overline e_{\rm C})$, using unrounded means.

\begin{table}[h]
\centering
\caption{All 27 existing runs: final-epoch test relative $L^2$, lower is better. Each row contains three separately trained models.}
\label{tab:seed_results}
\input{generated/seed_results}
\end{table}

\paragraph{Text pathway.}
The encoder uses 128-dimensional token embeddings, masked mean pooling, a biased $128\!\to\!256\!\to\!128$ MLP with GELU, and $L^2$ normalization. It is randomly initialized and jointly trained, with no language pretraining, attention, positional encoding, or dropout. Pooling excludes padding and is invariant to token order. The lowercase tokenizer retains alphanumeric characters, underscores, periods, plus/minus signs, and slashes. Each run fits its vocabulary on at most 4,096 training examples and pads or truncates to 128 tokens; unseen tokens share one unknown token. Vocabulary sizes, including padding and unknown tokens, are 11 for constant, 32 for aligned 2D, and 37 for aligned CNS3D. The output conditions each FNO block through $(1+\gamma)h+\beta$, with the linear map producing $\gamma,\beta$ initialized to zero.

\paragraph{Optimization and alignment.}
The encoder, FiLM heads, and FNO are optimized jointly with normalized-field mean squared error; aligned runs additionally backpropagate the weighted contrastive loss through the text encoder and trainable alignment heads. All use AdamW with learning rate $5\times10^{-4}$, weight decay $10^{-5}$, model gradient clipping at norm 1, and mixed precision disabled. The evidence encoder concatenates per-channel means, population standard deviations, minima, and maxima of normalized inputs and targets, then applies an MLP with hidden width 256, GELU, and output width 128. Biased linear $128\!\to\!128$ text and data projections precede normalization and InfoNCE with fixed temperature $0.07$. Final-epoch losses are approximately $1.386294$ in 2D and $0.693167$--$0.693221$ in 3D. Neither convergence near $\log B$ nor Proposition~1 identifies the cause of predictive-error differences.

\paragraph{Parameter accounting.}
Surrogate parameter counts are 4,743,937 / 4,877,313 / 4,880,001 in 2D and 18,889,397 / 19,006,261 / 19,009,589 in 3D for unconditioned / constant / aligned, using the same implementation counting convention. Vocabulary sizes explain the difference between conditioned models. The training-only evidence encoder and projections add 68,224 parameters in 2D and 76,416 in 3D.

\paragraph{Interventions and reproducibility.}
Interventions reuse the final aligned model and training vocabulary. Shuffling permutes identical descriptions; missing/constant substitutes the generic sentence. Corruption replaces task, dimension, channels, target step, and other metadata values. Paraphrasing changes the template while retaining facts. Numeric metadata serializes labeled facts as text, not direct numerical conditioning. Figure~\ref{fig:prompt_controls} summarizes $100(e_{\rm intervention}-e_{\rm correct})/e_{\rm correct}$ within each model across seeds. Tables are generated by \texttt{scripts/build\_review\_tables.py}; \texttt{generated/review\_evidence.json} records source metrics and hashes.

%% file: generated/seed_results.tex
\begin{tabular}{lrccc}
\toprule
Task & Seed & Unconditioned & Constant & Fixed prompt + align. \\
\midrule
Darcy2D & 7 & 0.032205 & 0.030741 & 0.049860 \\
 & 17 & 0.033520 & 0.030068 & 0.034352 \\
 & 27 & 0.045690 & 0.029452 & 0.032868 \\
\addlinespace
ShallowWater2D & 7 & 0.00155104 & 0.00118240 & 0.00135158 \\
 & 17 & 0.00155618 & 0.00129879 & 0.00128726 \\
 & 27 & 0.00197349 & 0.00170078 & 0.00148120 \\
\addlinespace
CNS3D & 7 & 0.409692 & 0.413856 & 0.391852 \\
 & 17 & 0.403565 & 0.398259 & 0.401415 \\
 & 27 & 0.400125 & 0.408121 & 0.420486 \\
\bottomrule
\end{tabular}